\documentclass[preprint,12pt,authoryear]{elsarticle}

\usepackage{amssymb}
\usepackage{amsmath}

\usepackage{url} 
\usepackage{tabularx}
\usepackage{array}
\usepackage{algorithm}
\usepackage{algorithmic}
\usepackage{caption} 
\usepackage{xcolor}
\journal{High-Confidence Computing}

\begin{document}

\begin{frontmatter}



\title{Hierarchical Federated Transfer Learning in Digital Twin-Based Vehicular Networks} 


\author{Qasim Zia\textsuperscript{a}, Saide Zhu\textsuperscript{b}, Haoxin Wang\textsuperscript{a}, Zafar Iqbal\textsuperscript{a}, Yingshu Li\textsuperscript{a}} 

\affiliation{organization={ Department of Computer Science, Georgia State University},
            addressline={7th Floor, 25 Park Place Bldg}, 
            city={Atlanta},
            postcode={30303}, 
            state={GA},
            country={USA}}
\affiliation{organization={ College of Information Sciences and Technology, Penn State Berks},
            addressline={288 N. Burrowes Rd}, 
            city={Reading},
            postcode={19610}, 
            state={PA},
            country={USA}}

\begin{abstract}
In recent research on the Digital Twin-based Vehicular Ad hoc Network(DT-VANET), Federated Learning (FL) has shown its ability to provide data privacy. However, Federated learning struggles to adequately train a global model when confronted with data heterogeneity and data sparsity among vehicles, which ensure suboptimal accuracy in making precise predictions for different vehicle types. To address these challenges, this paper combines Federated Transfer Learning (FTL) to conduct vehicle clustering related to types of vehicles and proposes a novel Hierarchical Federated Transfer Learning (HFTL). We construct a framework for DT-VANET, along with two algorithms designed for cloud server model updates and intra-cluster federated transfer learning, to improve the accuracy of the global model. In addition, we developed a data quality score-based mechanism to prevent the global model from being affected by malicious vehicles. Lastly, detailed experiments on real-world datasets are conducted, considering different performance metrics that verify the effectiveness and efficiency of our algorithm. 
\end{abstract}

\begin{graphicalabstract}
\includegraphics{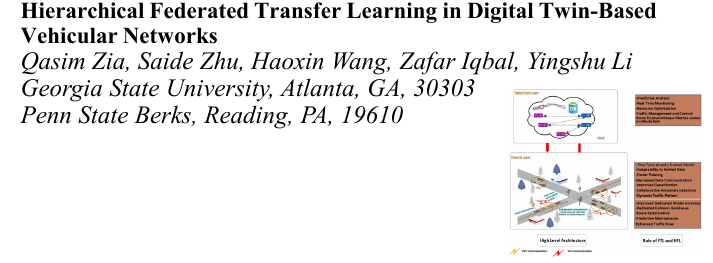}
\end{graphicalabstract}

\begin{highlights}
\item Research highlight 1
\item Research highlight 2
\end{highlights}


\end{frontmatter}



\section{Introduction}
\label{sec1}

Vehicular networks enable numerous applications for Intelligent Transportation Systems (ITSs), including traffic control, infotainment, and accident reporting \cite{noor2020survey}. These applications rely on various performance criteria, such as reliability, latency, and user-defined features like the quality of physical experience. Thus, the effective handling of vehicle heterogeneity, data privacy, and limited computational resources has become an urgent challenge to discuss, while current wireless technologies struggle to assist it thoroughly \cite{zia2016improving}. Despite advanced onboard sensors and processing capabilities of autonomous vehicles, they still face challenges such as limited sensing range, limited local processing capacity, and communication barriers \cite{he2022security}. To meet these diverse requirements, the development of vehicular networks must incorporate two key advancements: proactive, intelligent analytics and self-sustaining wireless systems.
\begin{table}[!t]
\caption{\textbf{Nomenclature}\label{tab:table1}}
\centering
\begin{tabular}{|c||c|}
\hline
\textbf{Symbol} & \textbf{Description}\\
\hline
$C_{n}$ & The Cluster’s Collection that are not trained\\
\hline
$i$ &$i^{th}$ Cluster \\
\hline
$v$ &The vehicle $v$\\
\hline
$j$ & The $j^{th}$ model update and the total number is equal to $| C_n |$\\
\hline
$n$ & Total number of clusters\\
\hline
$V_h, V_r, V_e$ & In a cluster, the head vehicle, route vehicle, and edge vehicle DT\\
\hline
$P_O$  & All the post vehicles that exist for any vehicle v in a given cluster\\
\hline
$DtQ$ & The data quantity that is used to train any given vehicle\\
\hline
$w^-$ &The model parameters that are used to send for training \\
\hline
$w$ & The model parameters that are received after training \\
\hline
$S_r$ & The score calculated for the credibility of the vehicle \\
\hline
$Data$ & The personal data of any vehicle \\
\hline
$W$ & The Global model parameters  \\
\hline
$VDT$ & Vehicle Digital Twin  \\
\hline
$DT$ & Digital Twin  \\
\hline
\end{tabular}
\end{table}

Proactive Intelligent Analytics is a data analytics technique that analyzes incoming data while anticipating potential challenges, trends, and opportunities in advance. It involves real-time or nearly real-time data analysis using the most advanced machine learning, artificial intelligence, and prediction models. Proactive online learning wireless systems will make it possible to optimize wireless resources to ensure the quality of service (QoS) of diverse ITS applications with varying requirements.  

Self-sustaining wireless systems can function without constant external power sources. These systems are helpful for Internet of Things(IoT) devices, remote locations, and scenarios where battery replacement or maintenance is challenging. Self-sustaining wireless networks will allow ITS to run with the least operator or user interaction. 

 The combination of digital twins and vehicular networks has become a promising paradigm in recent years, transforming our understanding of and approach to managing transportation systems. The development and growth of the digital twin provided insight into how to address the problems associated with autonomous vehicles. Virtual representation of physical systems serves as the foundation for digital twins. A digital twin is an accurate virtual representation of a physical thing on a cloud that shows its lifetime and status. The term “Digital Twin-based Vehicular Ad-Hoc Network” refers to the cloud-based digital twin of a physical vehicle that synchronizes real-time sensing data from actual vehicles through wireless communication. Although digital twins can enhance the understanding of vehicle systems through virtual representations, they still require efficient data processing and analysis technologies to enable decision-making. Distributed or centralized training can serve as a foundation for Machine Learning (ML) because vehicles are reluctant to share their private information, and centralized training causes privacy leaks by moving data from dispersed devices to a centralized location. At the same time, traditional centralized machine learning methods have proven insufficient when dealing with nonindependent and identically distributed (non-IID) data.

Artificial Intelligence (AI) and Machine Learning (ML) have revolutionized biomedical research, enabling breakthroughs in areas such as antibody design~\cite{ahmed2026chimerabench}, medical imaging~\cite{ahmed2023robust}, and protein-protein interaction prediction~\cite{ahmed2026epiformer}. This demonstrates the power of AI in understanding complex molecular interactions \cite{ahmed2025wallepp}. However, the data sharing remains a bottleneck for research due to privacy concerns. Federated Learning(FL) is the most recent technique in digital twin-based vehicular ad hoc networks that addresses centralized learning (CL) issues\cite{khan2023federated}. FL is well known for its ability to promote collaboration and protect data privacy in various domains, including industrial and medical cyber-physical systems \cite{abdulrahman2020survey}\cite{li2020deepfed}. FL, which employs dispersed vehicles to learn a global FL model without transporting data from devices to a centralized site for training, is presented as a solution to this centralized ML constraint \cite{khan2021dispersed}. Compared to centralized ML, FL maintains better privacy. However, traditional federated learning faces issues such as slow convergence and insufficient accuracy when dealing with the diversity of vehicle types and limited vehicle data. Therefore, we use the Hierarchical Federated Transfer Learning (HFTL) technique in digital twin-based vehicular networks (DT-VANET). HFTL groups vehicles into different clusters based on their type, and pre-trained models are used to perform personalized model fine-tuning that fits each vehicle type's specific needs. It also improves both prediction accuracy and convergence speed in a meaningful way.  
\begin{figure*}[!ht]
    \centering
    \includegraphics[width= \textwidth]{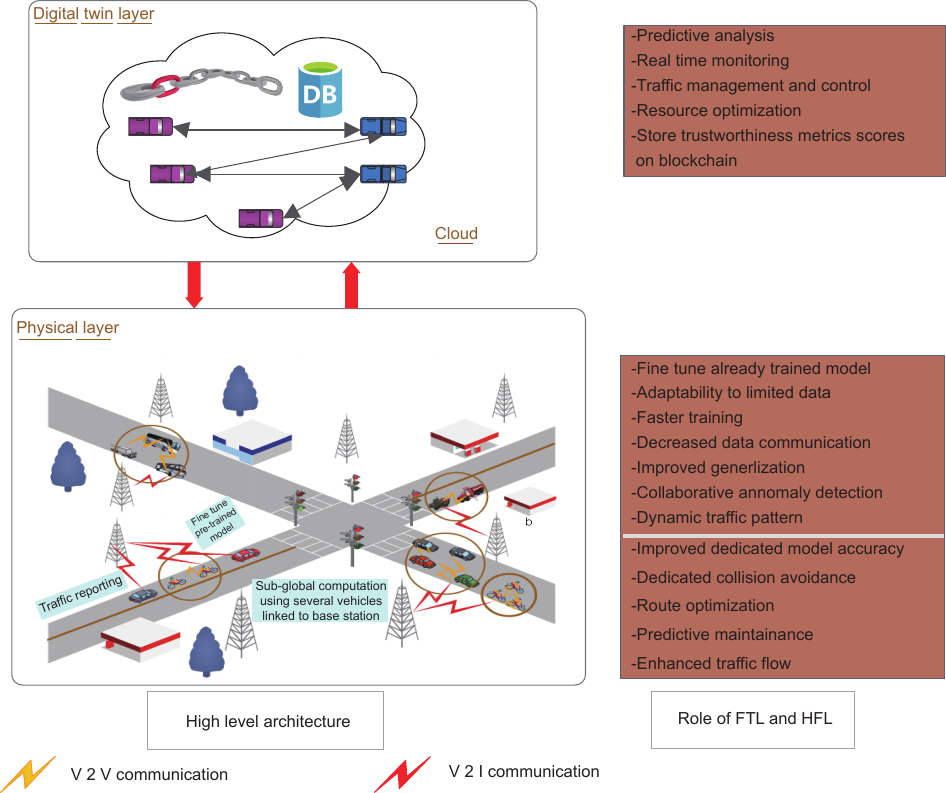}
    \caption{High-level architecture of Hierarchical Federated Transfer Learning for digital twin-based vehicular network}
    \label{fig_1}
\end{figure*}

We also resolved another issue related to optimizing the route associated with the specific vehicle type and managing the aggregation of communication overhead between various vehicle types that makes precise and accurate predictions for each vehicle type. Our HFTL for DT-VANET has also successfully resolved this issue. To ensure our architecture is trustworthy and reliable, each vehicle will assess its data quality, vehicle health, and safe driving performance score, and these scores will be stored in the blockchain of the cloud, which can be used to give rewards later.
We use blockchain's tamper-proof feature to ensure that recorded data cannot be maliciously altered. In that way, they are stopping untrustworthy vehicle nodes from uploading low-quality or fake data. Additionally, the distributed consensus mechanism of blockchain allows data quality scores to be shared across multiple nodes, making the global model updates more fair and reliable. Furthermore, the reward mechanism for vehicles can be automatically carried out through smart contracts.
So, our research paper addresses issues and challenges related to privacy, efficiency, accuracy, and scalability of machine learning models in digital twin vehicle networks, enhancing model performance through implementing an HFTL approach.   

To the best of our knowledge and understanding, our work is the first to review HFTL and Trustworthiness metrics in a single architecture of DT-VANET. Also, we illustrate a scenario to enable vehicle networks based on digital twins through the diagram. We also discussed the related work, algorithms of HFTL, and performance metrics in twin-based vehicular networks. Therefore, the following is a summary of our contributions:
\begin{enumerate}
\item We build a general thorough architecture for vehicular networks based on digital twins. A scenario for a digital twin-based vehicular network is shown, which enables efficient data synchronization between physical and virtual nodes of vehicles, enhancing the model's scalability and response speed. Additionally, we offer diagrams for twin-based vehicle networks. 
\item We outline the algorithms of federated transfer learning in digital twin-based vehicular networks. It allows for model customization based on different vehicle types, which improves prediction accuracy and training efficiency.

\item Lastly, we conducted experiments on real-time datasets to evaluate the performance metrics. By comparing the performance of HFTL, Federated Learning (FL), and Centralized Learning (CL) in terms of model accuracy, training time, communication overhead, and resource consumption, we validated the advantages of HFTL in different scenarios.
\end{enumerate}
The rest of the paper is structured as follows:
The related works are reviewed in Section \ref{sec:related_work}. We suggest a Weighted Cloud Server Cycling Model Update Algorithm and an Inner Cluster Federated Transfer Learning Algorithm in Section \ref{sec:ftl_based_comm}. Subsequently, in Section \ref{sec: eval}, comprehensive experiments are conducted and evaluated to determine the efficiency and effectiveness of our suggested architecture. Finally, Section \ref{sec:concl} discusses our conclusions.

\section{Related Works} \label{sec:related_work}
This section overviews recent research on Hierarchical Federated Learning, Federated Transfer Learning, and Clustering Techniques for VANET.
\subsection{Hierarchical Federated Learning in VANET}
Numerous research has presented the use of Federated Learning to address data privacy. However, Google suggested the first work on Federated Learning (FL) in 2016 \cite{konevcny2016federated} for conventional FL. In 2022, Li et al. \cite{li2022feel} proposed a FEEL framework using Federated Learning for non-IID data. To predict the high accuracy of network traffic with a high volume of data, Sepasgozar et al. \cite{sepasgozar2022fed} proposed a new Network Traffic Prediction (Fed-NTP) based on federated learning. Hierarchical Learning in Machine Learning was first introduced by Zhang et al. \cite{zhang2006hierarchical}. The human approach to problem-solving in hierarchies inspires their work. In 2021, Goncalves et al. \cite{gonccalves2021intelligent} discuss the Hierarchical approach for Security Framework in VANET. In 2024, HaghighiFard et al. \cite{haghighifard2024hierarchical} discuss the use of Hierarchical Federated Learning in VANET. It uses the cosine similarity of FL model parameters and average relative speed as a metric for making clusters.
Although Li et al.\cite{li2022feel}, the FEEL framework successfully applied federated learning to non-IID data environments. It did not fully address the data communication challenges in highly dynamic VANET environments. In contrast, HaghighiFard et al.\cite{haghighifard2024hierarchical} introduced hierarchical federated learning, which reduced communication overhead to some extent. However, their clustering method still relies on fixed vehicle attributes, making it less adaptable to the rapid changes in vehicle environments.

\subsection{Federated Transfer Learning in VANET}
Federated Transfer Learning(FTL) facilitates knowledge transfer without affecting user privacy. The FTL was first discussed by Liu et al. in 2020 \cite{liu2020secure}, where they enabled the target-domain party to use the source domain’s rich level to create adaptable and efficient models. In 2022, Otoum et al. \cite{otoum2022feasibility} proposed an intrusion detection system in VANET. On that account, they have compared Split Learning, Federated Learning, and Transfer Learning. However, Liu et al. \cite{liu2020secure} paper discusses Federated Transfer Learning. Still, the paper does not consider DT-VANET for their findings. Otoum et al. \cite{otoum2022feasibility} consider federated transfer learning only for intrusion detection scenarios, not for digital twin-based communication. 
\subsection{Clustering Techniques for VANET}
The primary purpose of clustering is to find discrete groups of vehicles in the environment. It will reduce the communication with the roadside units. \cite{zia2015survey} discusses various data-centric protocols related to clustering. The PasCar routing protocol was presented by Wang et al. \cite{wang2013passcar} in 2013 to make a reliable cluster structure based on a single hop. It chooses suitable participants from candidate nodes. However, the single-hop mechanism limits the system’s coverage and stability. Because of this, there has been a lot of research on multi-hop clustering techniques in the years that followed. In 2022, Rashid et al. \cite{rashid2020prediction} discussed efficient multi-hop clustering based on prediction. This paper discusses increasing the cluster coverage area increase, link stability, energy efficiency, and mobility characteristics increase. Temurnikar et al. \cite{temurnikar2022pso} in 2022 discussed a new Particle Swarm Optimization (PSO) based multi-hop method because of which you can use the best route. We can also select the Cluster head, which is much more stable. It also identified false messages from malicious vehicles. \cite{zia2024priority} in 2024, discussed the possibility of using priorities to achieve faster response time. 
Compared to the techniques already mentioned, this paper introduces Hierarchical Federated Transfer Learning (HFTL), which offers more flexible clustering to address vehicle heterogeneity. Additionally, integrating digital twin technology enables real-time data synchronization, improving prediction accuracy and reducing communication overhead.

\section{THE PROPOSED HIERARCHICAL FEDERATED TRANSFER LEARNING IN DIGITAL TWIN-BASED VEHICULAR NETWORKS}\label{sec:ftl_based_comm}
In this section, we proposed a high-level secure and robust federated transfer learning architecture for vehicular networks, as shown in Figure \ref{fig_1}. There is a digital twin layer, and the physical layer forms the two main layers of the proposed architecture. The physical layer consists of objects like base stations (BS), which help end users and distribute autonomous vehicles. The Digital Twin Layer consists of four essential parts: data storage, twin management, virtual model mapping, and blockchain \cite{dai2022adaptive}. It uses digital twins to effectively map the virtual twin system objects and real-vehicle edge computing. The cloud server implements the digital twin layer, and the idea of digital twin objects is used \cite{khan2022digital}. A virtual representation of the physical system is known as a digital twin object, and it is doable to simulate numerous vehicle network operations and applications using this virtual representation type of the physical vehicular network.  Mathematical and experimental methodologies can be used to model such network applications/functions\cite{khan2022digital2}. However, as we know, mathematical modeling is highly dependent on assumptions; it might not be able to denote the actual network accurately.
Meanwhile, mistakes may prevent experimental modeling from being accurate during experimentation. In digital twin-based vehicular networks, a data-driven modeling approach based on Federated Transfer Learning may be considered to overcome these problems. It also provides enhanced network performance and efficiency, improved safety and reliability, and allows advanced applications to run. Hierarchical Federated Transfer Learning(HFTL) in digital twins is used for precise and accurate prediction according to the vehicle type. Blockchain is also used to secure the trustworthiness metrics of the individual vehicle. We will now discuss the features of this architecture one by one. 
\begin{table*}[!t]
\small 
\caption{\textbf{Categorization of Vehicles}\label{tab:table2}}
\centering
\begin{tabularx}{\textwidth}{|c||c||X|}
\hline
\textbf{Category ID} & \textbf{Category}& \textbf{Types of Vehicles}\\
\hline
1 & Regular Vehicles & Cars\\
\hline
2 & Emergency Vehicles & Ambulances, Fire Trucks, Police Cars\\
\hline
3 & Public Transportation Vehicles & Buses\\
\hline
4 & Delivery Vehicles & Goods and Packages Transportation Vehicles \\
\hline
5 & Specialized Vehicles & Construction Vehicles, Utility Vehicles, Road Maintenance Vehicles\\
\hline
6 & Two-Wheel Regular Vehicles & Motorcycles\\
\hline
\end{tabularx}
\end{table*}

\subsection{Hierarchical Federated Transfer Learning (HFTL)}
HFTL's clustering considers vehicle size, purpose, driving patterns, and computational resources. It will help to build a customized model. This model aligns with the learning goals and data properties specific to each kind of vehicle. As shown in Figure \ref{fig_1}, vehicles are grouped according to the type of vehicles. The types of vehicles are classified according to vehicle size, purpose, and computational resources in Table \ref{tab:table2}. Generally, the vehicle with the most advanced features is selected as the cluster head. 
\begin{equation}\label{eq: H_cluster}
DT = \sum_{i = 1}^{n} \sum_{j = 1}^{n}  Tv_{j}\in  C_i
\end{equation}

Where $C_i$ is a cluster that contains a type of vehicle j.

Equation \ref{eq: H_cluster} discusses that in the Digital Twin, there are many clusters from $C_i$ to $C_n$ where each cluster $C_i$ contains only one type of vehicle $Tv_j$.

Secondly, HFTL improves the convergence and heterogeneity in Dispersed Federated Learning. In HFTL, a pre-trained model is shared depending on the type of vehicle and target application, such as traffic flow prediction, collision risk prediction, intelligent parking prediction, etc. This pre-trained model will be fine-tuned using the vehicle's private local data. This enables the model to be customized for the environment and based on the vehicle's driving patterns.
For example, public vehicles that require higher precision will have their model fine-tuning focus more on safety and route optimization. However, private vehicles may prioritize reducing idle times and optimizing the driving experience.
After training, share the model update to the cloud server for the global model update. Therefore, model updates are combined from every vehicle's digital twin to the cloud to make it a more robust and heterogeneous global model. Several digital twins of vehicles use these model updates to make decisions and predictions. The model updates from every vehicle digital twin are given weight according to the trustworthiness metrics. These metrics are recorded and verified through blockchain. Vehicles with higher trustworthiness have greater weight in global model updates, ensuring the security and accuracy of the model updates. These weights will be considered for the Global model update using the Federated Averaging Method. 
\begin{figure*}[!ht]
    \centering
    \includegraphics[width=\textwidth]{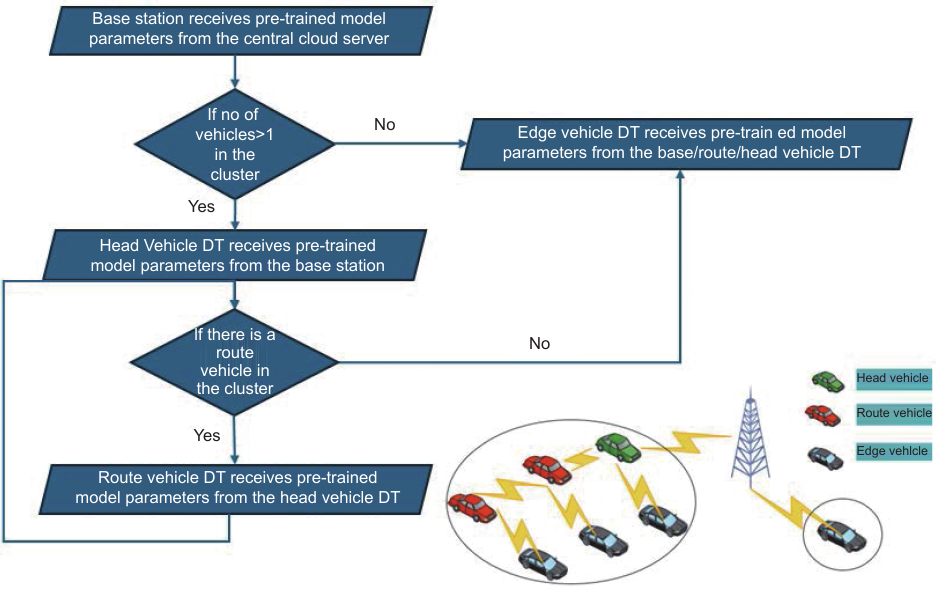}
    \caption{The Flowchart and Vehicle type data flow for the Inner Cluster of Hierarchical Federated Transfer Learning}
    \label{fig_2}
\end{figure*}
\subsection{Digital Twin Based Vehicular Networks}
As we discussed briefly above, the digital twin architecture in VANET allows us to have many benefits, including improved decision-making, increased safety, predictive maintenance, optimized network performance, enhanced scalability, enhanced simulation and testing, real-time monitoring and maintenance, and cost reduction.

\subsection{Blockchain Integration in Architecture}
In this architecture, blockchain can be defined as a distributed ledger integrated into DT-VANET-based HFTL architecture to manage and store weights. The trustworthiness metrics calculate every weight, as discussed in the next section. Low-weight model updates are not considered to update the global model to maintain integrity. It will also help us detect anomalies. We ensure the secure DT-VANET-based HFTL architecture through the weights and tamper-proof blockchain.
\subsection{Trustworthiness Metrics}
\begin{equation}\label{eq:H_trust}
\begin{split}
DTS_z = & \, w_1\cdot DTC_z + w_2\cdot SRC_z +  w_3\cdot E_z \\
& + w_4\cdot CSF_z + w_5\cdot GRHL_z + w_6\cdot SFR_z
\end{split}
\end{equation}
where: \\
\begin{itemize}
\item $DTS_z$ = The Data Quality Score of vehicle z \\
\item $DTC_z$ = The Data Completeness of vehicle z \\
\item $SRC_z$ = The sensor collaboration of vehicle z \\
\item $E_z$ = The events (traffic incidents, road conditions, etc.) recorded by vehicle z \\
\item $CSF_z$ = The constant format (According to Standard Data Formats) for vehicle z \\
\item $GRHL_z$ = The General Health of vehicle z (e.g., looking after maintenance issues)\\
\item $SFR_z$ = The reliable and safe driving patterns for vehicle z.
\item $w_1$,$w_2$,$w_3$...$w_6$ are the weights that are assigned with each factor
\end{itemize}
Every vehicle will calculate its data quality score depending on data completeness, sensor collaboration, events, constant format, the vehicle’s general health, and safe, reliable driving patterns. 
\begin{equation}
RPS_z(tm+1) = RPS_z(tm) + \alpha \cdot DQS_z - \beta \cdot M_z
\end{equation}
where: \\
\begin{itemize}
\item $RPS_z$(tm) = The Reputation Score of vehicle z \\
\item $\alpha$ = The Weight for data quality score \\
\item $\beta$ = The malicious behaviour Penalty factor \\
\item $M_z$ = The malicious activity indicator for vehicle z (0 or 1) \\
\item $RPS_z$(tm + 1) = At tm + 1, updated reputation score of vehicle z
\end{itemize}
\begin{equation}
Wt_z = \frac{RPS_z}{\sum_{m=1}^{n} RPS_m}
\end{equation}
where: \\
\begin{itemize}
\item $WT_z$ = Weight of vehicle z's model update.
\item $n$ = Total number of vehicles
\end{itemize}
These scores are input to the reputation system maintained at the cloud end. This reputation system helps DT-VANET to assign weight to the model update as described in the previous section. It will also help to keep the malicious vehicle from interfering. 
\begin{equation}
\Delta RPS_z = RPS_z(tm+1) - RS_z(tm)
\end{equation}
\[
BC_z = \{ \Delta RPS_z(tm_1), \Delta RPS_z(tm_2), ..., \Delta RPS_z(tm_n) \}
\]
where: \\
\begin{itemize}
\item $BC_z$ is the set of reputation changes over the time {$t_1,t_2,...t_m$} for vehicle z  
\end{itemize}
The reputation of a vehicle changes over time, and digital twins on the cloud are in charge of figuring out and recording these value changes on the blockchain.
\begin{equation}
RW_z = f(RPS_z) \quad \text{where} \quad f'(RPS_z) > 0
\end{equation}
Where:
\begin{itemize}
\item $RW_z$ for vehicle z be a function of reputation score $RPS_z$
\end{itemize}
\[
RW_z = \gamma \cdot RPS_z
\]
Where:
\begin{itemize}
\item $\gamma$ is used for changing reputation score $RPS_z$ into a specific reward with a scaling factor.
\end{itemize}
Vehicles with good scores and positive contributions may be rewarded with tax or toll reductions or urgent access to services.
\begin{equation}
RPS_z < \theta \implies \text{Vehicle } z \text{ is excluded from participation.}
\end{equation}
Where:
\begin{itemize}
\item Malicious Vehicle reputation score $RPS_z$ for vehicle z is considered dangerous and stopped from participation  
\end{itemize}
A blockchain ensures tamper-proof records of participation, high-quality data, and protocol enforcement.

\subsection{Cloud Server Hierarchical Federated Transfer Learning}
\begin{algorithm}[H]
\caption{Weighted Cloud Server Cycling Model Update}\label{alg:alg1}
\begin{algorithmic}[1]
\STATE The central cloud server has existing model parameters $W_{exist}$ and digital twin of the entire system for simulation;
\STATE Initialize $C_n$ ;
\FOR{ each update $j = 0,1,2, \cdots ,j - 1$ }
    \STATE$W_{exist}^-$ = $W_{exist}$;
    \STATE Select cluster i from $C_n$;
    \STATE Send $W_{exist}^-$  to Digital Twin of cluster i ;
    \STATE c  =\textbf{\textsc{Inner-Cluster}}($W_{exist}^-$ );
    \STATE Calculate $F_{weight}$($W_{exist}^+$) ;
    \STATE Eliminate i from $C_n$;
    \STATE Update $W_{exist}$ ;
\ENDFOR
\STATE Acquire the final model parameters $W_{exist}$ ;
\end{algorithmic}
\label{alg1}
\end{algorithm}

\subsubsection{Initialization}
First, the central cloud server has existing model parameters for the Federated Transfer Learning Process and the digital twins for the entire VANET. It initializes the collection of clusters that have not yet participated in Federated Transfer Learning.

\subsubsection{Selection of Training Clusters}
We have a collection of clusters $C_n$ that participates in FTL Learning. If we want to take updates from all the clusters in the environment, then $C_n$ = n. There are j model updates in total. Each cluster contains a group of specific types of vehicles. The central cloud server sends the latest model parameters to a base station that forwards it to the particular cluster in $C_n$ that needs the prediction to be done. The cluster may be chosen based on the requirement to make some predictions. Pre-trained model parameters at the start of the jth update are sent to the base station of a selected cluster's digital twin in $C_n$. This allows the vehicle in that cluster to carry out inner-cluster training to derive the model parameters $W_{\textbf{exist}}^\textbf{-}$. The details of the inner-cluster training will be covered in the following subsection. It is not observable to the central cloud server.

\subsubsection{Updates to the Cloud Central Model}
We denote the model parameters that the central cloud server sends to and receives from the base station for cluster i in the jth update as  $W_{\textbf{exist}}^\textbf{-}$ and $W_{\textbf{exist}}^\textbf{+}$. We calculate the weight of the received model parameters using the scores we received for each vehicle in the given cluster. 
\begin{equation}\label{eq: F_weight}
F_{\text{weight}}(W_{\text{exist}}^+) = \frac{\sum Sr}{|v|}
\end{equation}
In the above equation \ref{eq: F_weight}, $\sum Sr$ is the sum of all the scores that each vehicle DT calculated in the given cluster, and $|v|$ is the number representing all the vehicles in each cluster. If there is only one vehicle in the cluster due to a lack of homogeneous vehicle type in the surroundings, we will consider a cluster of only one vehicle. This overall provides us with the average score of vehicles in a cluster. However, If the output weight is below a certain threshold, regarded as the data quality it is trained on is not good enough to update the global model ModelUpdate(.) at the central cloud server. Cluster i will be removed from the $C_n$ after deciding whether to update the global model parameters based on the average score of clusters. After j updates, a complete network model with parameters $W_{exist}$ is determined. Algorithm \ref{alg:alg1} indicates that our suggested architecture framework needs j-cluster iterations of the inner-cluster algorithm for the central cloud server updates.

\subsection{Inter Cluster Hierarchical Federated Transfer Learning}
\begin{algorithm}[H]
\caption{Inner Cluster Hierarchical Federated Transfer Learning}\label{alg:alg2}
\textbf{Input:} The model parameters of the cluster's digital twin that need to be fine-tuned  $W_{\textbf{exist}}^\textbf{-}$; \\
\textbf{Output:} The updated fine-tuned trained model parameters   $W_{\textbf{exist}}^\textbf{+}$ ;
\begin{algorithmic}[1]
\STATE \textbf{\textsc{Inner-Cluster}}():
\STATE \textbf{\underline{(I). For base station:}}
\STATE Receive $W_{exist}^-$ from the central cloud server;
\STATE Send $W_{exist}^-$  to $V_h$ ;
\STATE Receive $W_{exist}^+$ and Sr from $V_h$;
\STATE Send $W_{exist}^+$ and Sr to the central cloud server;
\STATE
\STATE \textbf{\underline{(II). For head vehicle DT $V_h$:}}
\STATE Receive $w_{h}^-$ = $W_{exist}^-$  from the base station;
\STATE Note its post arrange vehicles into set $Po_h$;
\FOR{ all vehicles $v \in Po_h$ }
    \STATE Send $w_{h}^-$  to vehicle DT ;
    \STATE Receive $w_{v}$ , $DQ_v$ and Sr from each vehicle DT;
    \STATE Update $w_{h}^-$ = combine($w_{h}^-$ , $w_{v}$ , $DtQ_v$ , Sr) ;
\ENDFOR
\STATE $w_{h}^-$ = \textbf{\textsc{Fine\_Tun}}($w_{h}$ , $DtQ_v$) ;
\STATE Send $w_{h}^+$ and Sr to the base station;
\STATE
\STATE \textbf{\underline{(III). For route vehicle DT $V_r$:}}
\STATE Receive $w_{r}^-$  from its previous vehicle DT  ;
\STATE Note its post arrange vehicles into set $Po_r$;
\FOR{ all vehicles $v \in Po_r$ }
    \STATE Send $w_{r}^-$  to vehicle DT ;
    \STATE Receive $w_{v}$ , $DQ_v$ and Sr from each vehicle DT;
    \STATE Update $w_{r}^-$ = combine($w_{r}^-$ , $w_{v}$ , $DtQ_v$ , Sr) ;
\ENDFOR
\STATE $w_{r}^-$ = \textbf{\textsc{Fine\_Tun}}($w_{r}$ , $DtQ_v$) ;
\STATE Update $DtQ_r$ and Sr;
\STATE Send $w_{r}$, $DtQ_r$ and Sr to its previous vehicle DT;
\STATE
\STATE \textbf{\underline{(IV). For edge vehicle DT $V_e$:}}
\STATE Receive $w_{e}^-$  from its previous vehicle DT;
\STATE $w_{e}^-$ = \textbf{\textsc{Fine\_Tun}}($w_{e}$ , $DtQ_e$) ;
\STATE $DtQ_e$ = $|Data_{ev}|$;
\STATE Send $w_{e}$, $DtQ_e$ and Sr to its previous vehicle DT;
\end{algorithmic}
\label{alg1}
\end{algorithm}

The Algorithm \ref{alg:alg2} sequence of collaboration and flowchart across vehicles in a cluster is depicted in Figure \ref{fig_2}.In our proposed framework, vehicles are further divided into three groups in a cluster named edge vehicle DT $V_e$, route vehicle DT $V_r$, and head vehicle DT $V_h$ if there is more than one vehicle in a cluster which draws inspiration from earlier work \cite{haghighifard2024hierarchical} that separates vehicles in each cluster into cluster head and cluster members. 

Assumption 1: The Inner Cluster training occurs on any cluster i digital twin that belongs to the $Cn$ cluster's collection. We assume that our cluster consists of head, route, and edge vehicles. However, if any cluster is not formed of the same vehicle collection, the relevant portion of the cluster will be discarded.
\subsubsection{Base Station}
After the head vehicle, DT $v_h$, completes its training, it shares the updated model parameters and data quality score with the base station. The Base Station then shares its model parameters and data quality score with the global model that resides on the cloud server. The relevant global model will be updated only after the cluster has a good aggregated data quality score.
\subsubsection{Head Vehicle}
When there is more than one vehicle in the cluster, there is always a Head Vehicle $v_h$. No vehicle exists before the head vehicle $v_h$ in the cluster. It is directly connected to the base station if there is more than one vehicle. For HFTL to execute, the base station sends global model parameters to $v_h$ vehicle DT, which forwards global parameters to the vehicle DT that exists next to this vehicle DT in the cluster. To represent the collection of all the vehicles that are next to $v_h$ vehicle, we use $Po_h$. The updated fine-tuned global parameters $w_{v}$, for constructing model parameters Data Quantity $DQ_v$ and Sr is the Data Quality scores are received by the $v_h$ vehicle DT for integration when the next vehicle DT is finished with their fine-tuning. 

While integrating, we give weightage to the parameters according to the amount of data on which it is fine-tuned and their data quality scores. After this, the $v_h$ vehicle DT forwards the updated integrated model parameters and scores to the next vehicle's DT, which remains in the cluster for fine-tuning. The above procedure continues until all the remaining vehicle DTs are completed with the fine-tuning. Ultimately, the model is fine-tuned using the local data of $v_h$ vehicle DT, and the Data quality score is calculated. The finalized fine-tuned model parameters and Data Quality Score are then sent to the base station.

\subsubsection{Route Vehicle}
A vehicle that functions as a routing vehicle DT is represented by the symbol $v_r$, which is a type of vehicle that has at least one predecessor vehicle and one successor vehicle. The $v_r$ vehicle DT gets the model parameters $w_{r}^-$ from $v_h$ the predecessor vehicle DT and sends them to all the successor vehicles DT one at a time. Like $v_h$ DT, $v_r$ DT integrates $w_{r}^-$ with model parameters received after fine-tuning from the successor vehicle and continues with the same process until the successor vehicles in the collection are done with the same. After that, $v_r$ DT, using its local data, fine-tunes the model parameters and calculates the Data Quality Score. After this, $v_r$ sent it to the predecessor vehicle.

\subsubsection{Edge Vehicle}
The term $v_e$ describes a vehicle DT that is either the only vehicle in the cluster or has only predecessor vehicles but no successor vehicles. If it is the only vehicle, it will forward model parameters directly to the base station; otherwise, it will send them to the predecessor vehicles. When $v_e$ DT gets model parameters from its predecessor vehicle's DTs, it performs fine-tuning and sends back the model parameters and data quality score(Sr) to the predecessor vehicle DT. 

By making clusters of similar vehicle types, we also make a low communication overhead with fewer links to the base station. We also make sure that our trained model is customized to each type of vehicle separately.
\begin{figure*}[!ht]
    \centering
    \includegraphics[width = \textwidth]{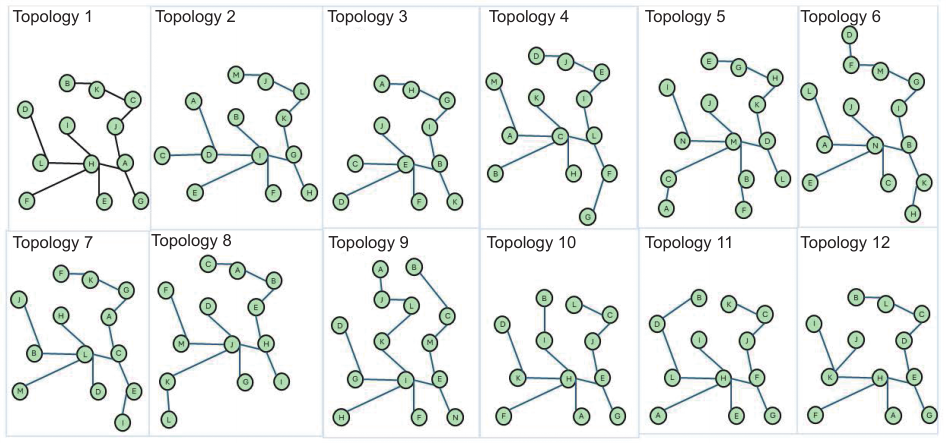}
    \caption{Twelve random topologies Network X generated}
    \label{fig_7}
\end{figure*}

\subsubsection{Example Scenario}
We present an example of specific Intelligent Transportation System(ITS) applications where Hierarchical Federated Transfer Learning (HFTL) in Digital Twin Vehicular Networks could be particularly beneficial:
\begin{enumerate}
\item Traffic Flow Management\\
HFTL in DT-VANET can enhance traffic flow by aggregating data from sensors of multiple vehicles to accurately predict real-time traffic congestion patterns. The digital twin also adjusts the signal timings in real-time according to traffic conditions, helping to maintain traffic flow.
\item Predictive Maintenance\\
By analyzing the patterns of component failures, the system can predict the maintenance needs of the vehicle's components(e.g., Brake issues, engine performance).
\item Real-time Route Optimization\\
The system can provide route recommendations based on the weather, traffic, and driver patterns. The digital twin can also optimize routes for emergency vehicles.
\item Safety Applications\\  
The system can also alert drivers of a potential collision by acquiring data from multiple vehicles. The architecture can also be beneficial for getting weather updates from vehicles and for providing predictions and alerts to upcoming vehicles about the weather.
\item Smart Parking, Charging and Environmental Solutions\\
Digital Twins helps vehicles find parking and charging spots in crowded areas. It also helps identify the polluted regions of the city.
\end{enumerate}
\section{PERFORMANCE EVALUATION}\label{sec: eval}
This section includes a thorough evaluation of our suggested framework using Hierarchical Federated Transfer Learning (HFTL) with Clustered Federated Learning (CFL), Federated Learning (FL), and Centralized Learning (CL) by different performance evaluations. We chose Centralized Learning (CL), Federated Learning (FL), and Clustered Federated Learning (CFL) as our baseline comparisons because CL represents the simplest centralized training method, while FL represents distributed learning. Conversely, CFL extends federated learning with clusters that resonate with our algorithm.  The above techniques conflict with HFTL regarding data heterogeneity and resource utilization, effectively highlighting the advantages of HFTL in heterogeneous vehicle environments. The generation of the cluster topology in our experiment is based on the Network X network structure, using a random number of vehicles distributed according to different vehicle types. There are twelve clusters, each having more than a random ten number of vehicles. It helps us to analyze results in high-density traffic scenarios, as shown in Figure \ref{fig_7}. It also helps to evaluate the system's ability to handle high-traffic data, which is very helpful for performance metrics results. Although the number of vehicles in each cluster is random, the clustering is based on vehicle type. To ensure the realism of the experiment in each generated cluster, only one head vehicle connects with the base station, and the rest of the vehicles are either routing vehicles or edge vehicles, simulating the actual distribution of different vehicle types. In Figure \ref{fig_7}, node A represents the head vehicle, and the rest of the node represents vehicles that are either route or edge vehicles. Below are the specific experimental settings.

\subsection{Experiment Settings}
\subsubsection{Experimental Setup}
Network X\footnote{A Network structure in Python (https://networkx.org/)}creates the random Inner Cluster topologies while Torch\footnote {Deep Learning Library for Python(https://pytorch.org/)} implements our suggested federated transfer learning framework. The Google Colab\cite{Googlecolab_2022} is used for implementation.

\subsubsection{Setting for Cluster}
We form twelve random clusters, each with a variable number of vehicles but the same vehicle type as described in table \ref{tab:table2}. As mentioned in Figure \ref{fig_2}, if each cluster has more than one vehicle, there would be a single cluster head; otherwise, in the case of the single vehicle, it would act as an edge vehicle. In our case for the evaluation of performance metrics, we keep the number of vehicles to more than a random ten number of vehicles because it will help us  to evaluate the system's ability to handle high traffic data as shown in Figure \ref{fig_7}
\subsubsection{Datasets}
For our performance metrics, we have conducted our experiments on the real-world dataset named vehicle mobility trace \footnote{https://vehicular-mobilitytrace.github.io/index.html\#data}. The data sets comprise real-time values, such as vehicle position, angle, vehicle ID, coordinates, speed, etc. As part of preprocessing, we first perform data cleaning, which handles outliers by replacing them with extreme values and using interpolation techniques to fill in missing data. After that, we use a standard scalar to normalize the data. It makes sure that all features are on the same scale. We split the data set in such a way that 80 \% is used as training while 20\% as testing. The main reason behind this data split is to make sure there is enough data for training to retain some portion for evaluation. However, we also used cross-validation to verify the model performance across several data subsets to guarantee the reliability and stability of the results. We assigned vehicles a numerical ID according to their category, as mentioned in Table \ref{tab:table2}. This will help us implement Hierarchical Federated Learning by randomly clustering vehicles according to their similar category. 

\subsubsection{Baseline Studies}
We will compare centralized learning, federated learning, clustered federated learning, and federated transfer learning. For comparison, we have used centralized, federated learning, and clustered federated learning as they are most closely related to federated transfer learning. Furthermore, while implementing the search for the most efficient hyperparameters, we utilized a checkpoint that led us to set the number of epochs to 200, the learning rate $(\alpha)$ to 0.1, and the batch size to 128 for both the proposed Federated transfer learning and the baseline algorithms.

\subsubsection{Performance Metrics}

We computed average Model Accuracy, Training Time, Resource Consumption (Computational), Communication Overhead, Latency, Convergence Time, and Throughput. Model accuracy tells us our model's accuracy in correctly predicting output. Training time refers to the duration required to complete the training process. Resource Consumption refers to the utilization of computational resources during the training process. Communication Overhead describes the amount of data exchanged during transmission. The amount of time required with an input and producing output (prediction) by the model is known as Latency. Convergence Time is the time needed for the training procedure to attain the required accuracy or stable condition (in our case, 90\% accuracy). The number of predictions a model can make in an amount of time given is known as Throughput.
\begin{figure*}[!ht]
    \centering
    \includegraphics[width=\textwidth]{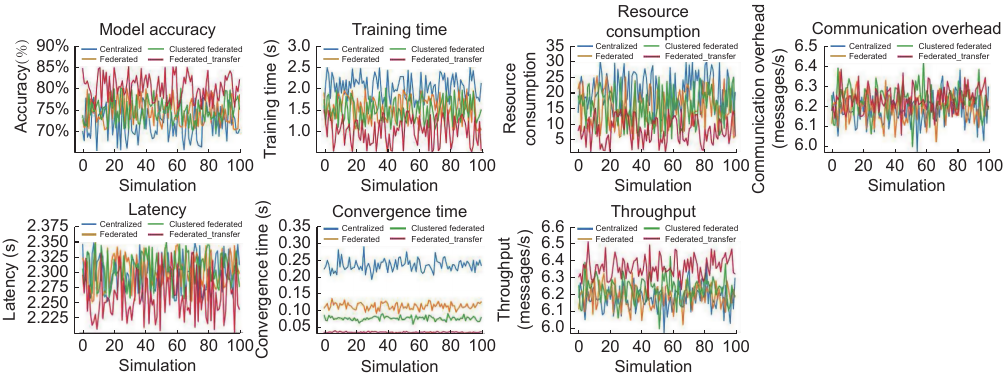}
    \caption{Performance metrics graph results of our proposed Federated Transfer Learning algorithm with Clustered Federated Learning, Federated Learning, and Centralized Learning}
    \label{fig_3}
\end{figure*}

\subsubsection{Experimental Results and Performance Analysis}
 We have compared existing centralized learning, federated learning, and clustered federated learning with federated transfer learning. We used a real-time vehicular mobility trace CSV file dataset to make our findings realistic. Figure \ref{fig_3} shows the graph for the performance metrics with 100 simulations to eliminate any bias scenario. Table \ref{tab:table3} shows the average performance of our HFTL algorithm over CL and FL. 
 \begin{table*}[!t]
\captionsetup{justification=centering}
\caption{\textbf{Experimental results regarding the average performance metrics across learning algorithms with Federated Transfer Learning Algorithm}\label{tab:table3}}
{\footnotesize 
\setlength{\tabcolsep}{1pt} 
\begin{tabularx}{\linewidth}{|c|*{7}{X|}}
\hline
\textbf{Algorithm} &\textbf{Model Accur} & \textbf{Train Time}& \textbf{Resour Consu}& \textbf{Comm Ohead}& \textbf{Latncy}& \textbf{Conver Time}& \textbf{Thrput}\\
\hline
Centralized Learning & 0.708& 2.03s& 20.19\%&6.19&2.323s&0.23s &6.20\\
\hline
Federated Learning & 0.721&1.53s&15.65\% &6.25&2.314s&0.12s&6.23 \\
\hline
Clustered Federated Learning & 0.752&1.46s&14.96\% &6.21&2.304s&0.08s&6.24 \\
\hline
Federated Transfer Learning & 0.822&1.02s&8.30\%& 6.28 & 2.240s &0.04s&6.35\\
\hline
\end{tabularx}
}
\end{table*}
 \footnote{Model Accur = Model Accuracy (Percentage \%), Train Time = Training Time (Sec), Resour Consu = Resource Consumption (Percentage \%), Comm Ohead = Communication Overhead (Messages), Latncy = Latency (Sec), Conver Time = Convergence Time (sec), Thrput = Throughput (Messages)}
 
The Federated transfer learning-based algorithm has the highest model accuracy average because it uses a pre-trained model according to the specific vehicle type. So, its prediction accuracy is better on average than the other two models.

 For the performance metric Training Time, on average, of a hundred different simulations, our algorithm takes less time because fine-tuning a model requires less time in seconds compared to training a model from scratch. 

 The federated transfer learning model requires fewer computational resources for resource consumption than the other models. That is because fine-tuning requires, on average, fewer computational resources than training from scratch, especially when we have customized models for each vehicle type.

The federated transfer learning algorithm has slightly more communication overhead when compared to Centralized learning(CL), Federated learning(FL), and Clustered Federated Learning (CFL). First, a pre-trained model must be fine-tuned and optimized based on each vehicle's personalized requirements, which will take up more communication overhead. Centralized learning(CL), Federated learning (FL), and Clustered Federated Learning (CFL) don't require pre-trained models for training, so their communication overhead is lower than our algorithm's. Having a slight difference in communication overhead does not affect our algorithm efficiency. However, the increased communication overhead is offset by significant improvements in accuracy and latency, giving HFTL a clear advantage in real-time decision-making and prediction precision, particularly in scenarios with high vehicle heterogeneity.

 The Federated Transfer Learning algorithm has less latency than Centralized learning(CL), Federated learning(FL), and Clustered Federated Learning (CFL). This is due to its use of pre-trained models that require less time to make predictions. Because of pre-trained model fine-tuning, this algorithm also has more throughput, faster convergence, a hierarchical structure, and low resource usage compared to Federated and Centralized learning. 
 \begin{figure*}[!ht]
    \centering
    \includegraphics[width=\textwidth]{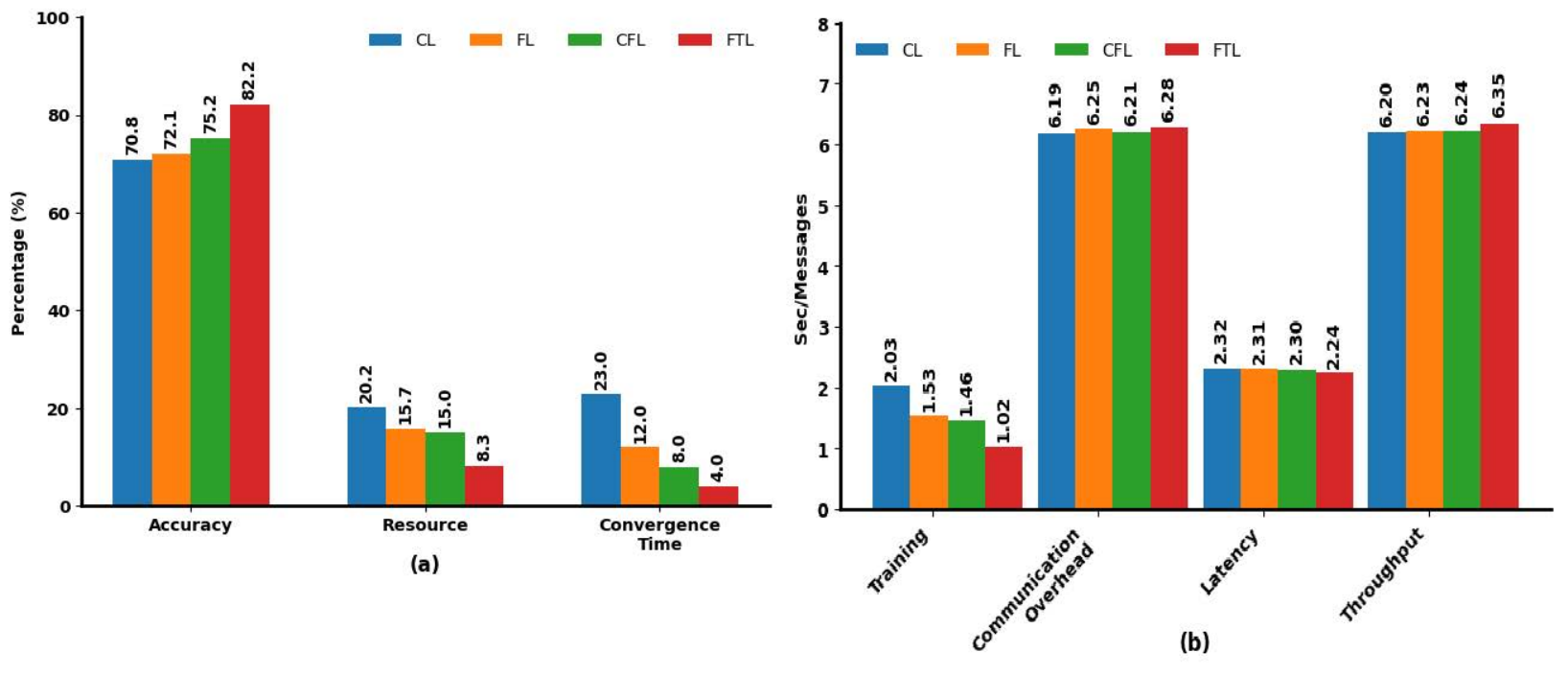}
    \caption{Graphical Comparison of Different Algorithm’s Model Accuracy, Resource Consumption, Convergence Time, Training Time, Communication Overhead, Latency, Throughput}
    \label{fig_4}
\end{figure*}

Figure \ref{fig_4} shows the bar graph comparison of HFTL over CFL, FL, and CL for Model Accuracy, Training Time, and Resource Consumption. HFTL outperforms the other algorithms. However, Figure \ref{fig_4} shows the bar graph comparison of HFTL over CFL, FL, and CL for Communication Overhead, where the other algorithm's communication overhead is slightly lower than HFTL.
\begin{figure*}[!ht]
    \centering
    \includegraphics[width=\textwidth]{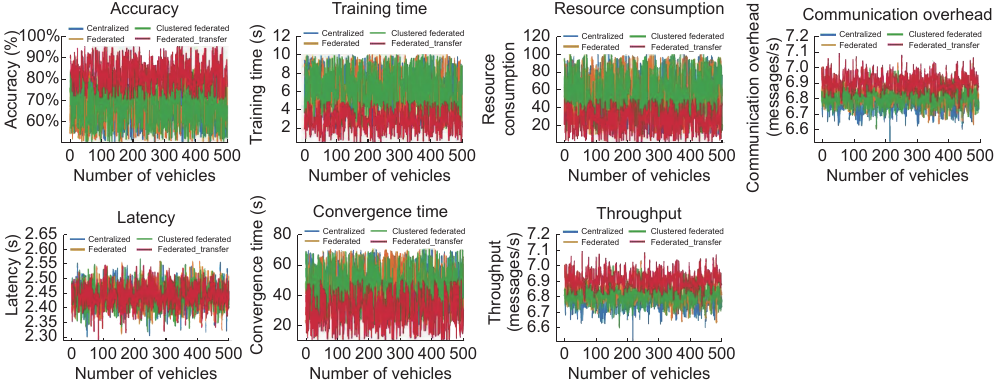}
    \caption{Scalability of our Federated Transfer Algorithm with Clustered Federated, Federated, and Centralized Learning}
    \label{fig_5}
\end{figure*}

\subsubsection{ Scalability Experimental Results and Performance Analysis}
The above results are tailored to a few vehicles and specific network conditions. We want to see how our algorithm performs if we gradually increase the number of vehicles in our network to check whether it remains effective.
Figure \ref{fig_5} shows the performance of learning algorithms for scalability metrics. In the Dynamic Environment, as compared to Clustered Federated, Federated, and Centralized Learning, our proposed algorithm performs better except for some underperformance in communication overhead. Table \ref{tab:table4} shows the average scalability metrics values for the algorithms.
\begin{table*}[!t]
\captionsetup{justification=centering}
\caption{\textbf{Experimental results regarding the average scalability metrics across the learning algorithms with Federated Transfer Learning Algorithm}\label{tab:table4}}
{\footnotesize 
\setlength{\tabcolsep}{1pt} 
\begin{tabularx}{\linewidth}{|c|*{7}{X|}}
\hline
\textbf{Algorithm} &\textbf{Model Accur} & \textbf{Train Time}& \textbf{Resour Consu}& \textbf{Comm Ohead}& \textbf{Latncy}& \textbf{Conver Time}& \textbf{Thrput}\\
\hline
Centralized Learning & 0.682& 6.01s& 54.06\%&6.77&2.47s&45.82s &6.79\\
\hline
Federated Learning & 0.689&6.07s&54.73\% &6.82&2.46s&45.72s&6.81 \\
\hline
Clustered Federated Learning & 0.68&5.96s&53.61\% &6.81&2.45s&45.46s&6.84 \\
\hline
Federated Transfer Learning & 0.823&2.78s&25.87\%& 6.90 & 2.42s &29.29s  &6.91\\
\hline
\end{tabularx}
}
\end{table*}

\begin{figure*}[!ht]
    \centering
    \includegraphics[width=\textwidth]{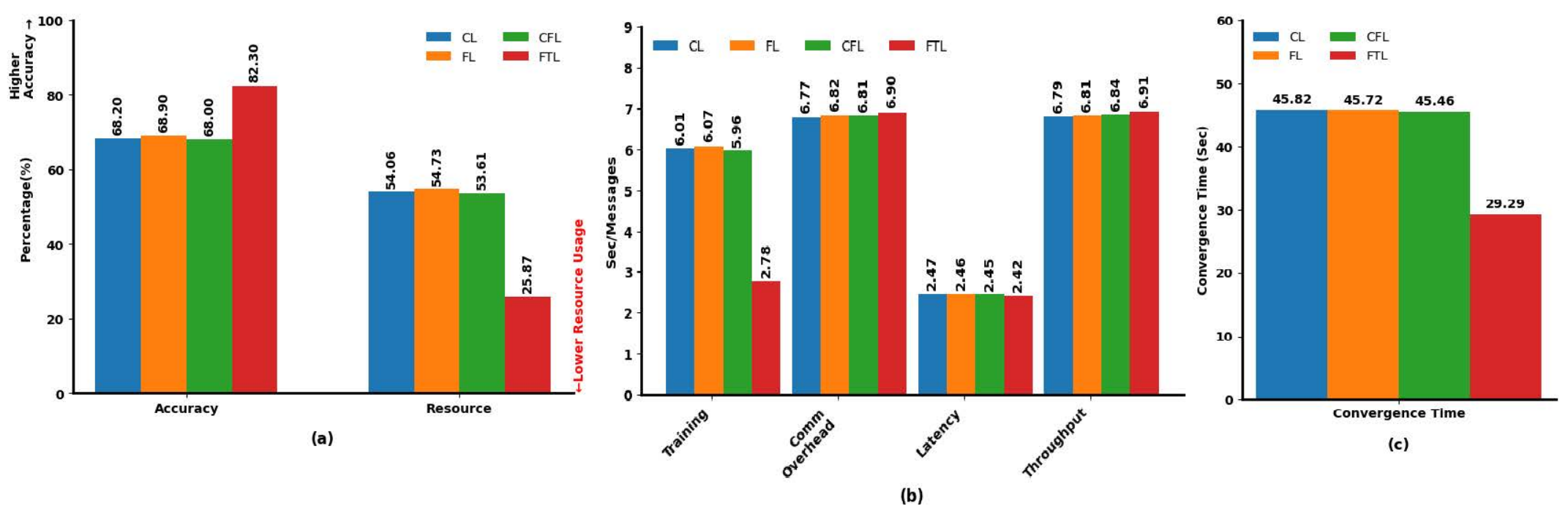 }
    \caption{Scalability of our Federated Transfer Algorithm with Federated and Centralized Learning}
    \label{fig_6}
\end{figure*}

Figure \ref{fig_6} shows these algorithms' differences in scalability performance metrics output as a bar graph.
\footnote{Model Accur = Model Accuracy (Percentage \%), Train Time = Training Time (Sec), Resour Consu = Resource Consumption (Percentage \%), Comm Ohead = Communication Overhead (Messages), Latncy = Latency (Sec), Conver Time = Convergence Time (sec), Thrput = Throughput (Messages)}

Scalability testing in large-scale VANETs shows that HFTL maintains high model accuracy and fast convergence as the number of nodes increases. However, as the network grows continuously, communication overhead and resource consumption may become limiting factors according to the results. Future work could explore communication compression techniques or asynchronous update mechanisms to further optimize HFTL's performance in large-scale networks.
As we increase the number of vehicles, the communication overhead increases based on the following:
\begin{enumerate}
\item  The model updates that need to be communicated also increase.
\item Aggregation complexity also increases communication overhead as vehicle updates are integrated with the head vehicle DT and sent to the central server. 
\item  More Communication Rounds are needed to achieve convergence because the number of vehicles grows, increasing communication overhead.
\end{enumerate}
As the number of vehicles increases, the communication overhead increases because of more data traffic and aggregation complexity. A slightly higher communication overhead than the other algorithms will not affect efficiency, as other performance metrics have greatly improved. 
\section{Challenges and Limitations}
Although the above results show the significance of our proposed research, some challenges and limitations are associated with it. 
\begin{enumerate}
\item  Data heterogeneity:
Significant differences in the data distribution between digital twins could result in less accuracy, more training time, and inefficient model convergence. 
\item  Network limitations:
Real-time synchronization and data transfer between vehicles and their respective digital twins can be affected by environments with erratic, unstable, or low-bandwidth communication links.
\item  Model Bias:\\
The global model may inherit bias if the local training data is biased
\item  Dynamic Contexts:\\
The hierarchical model may take time to adjust in real-time where the vehicle patterns change quickly
\item Heterogeneous VANET Types:\\
Our HFTL works well in urban and highways. It may struggle in sparely populated regions where data is less available. 
\end{enumerate} 
\section{Conclusion}\label{sec:concl}
This paper proposes the Hierarchical Federated Transfer Learning architecture and algorithm for the Digital Twin-based Vehicular Ad Hoc Network that provides faster and more accurate predictions. We have also formed clusters based on the specific vehicle type to make precise decision-making. We first designed the high-level architecture of Hierarchical Federated Learning. Then, we describe algorithms for our proposed architecture that are time-efficient and more accurate. Our experimental results demonstrate that HFTL outperforms Clustered Federated Learning (CFL), Federated Learning (FL), and Centralized Learning (CL) across several key metrics, including model convergence speed, throughput, and resource consumption. We also scale our network to see how our technique performs in the dynamic environment. In the future, we plan to conduct further research in the following areas: First, we aim to explore efficient communication compression techniques to reduce communication overhead. Second, we investigate asynchronous model update strategies to improve system real-time performance. Lastly, we intend to optimize lightweight blockchain implementations to ensure data privacy and security in large-scale vehicular networks.

\bibliographystyle{elsarticle-harv}

\end{document}